\documentclass[conference]{IEEEtran}
\IEEEoverridecommandlockouts

\usepackage{cite}
\usepackage{amsmath,amssymb,amsfonts}
\usepackage{algorithmic}
\usepackage{graphicx}
\usepackage{textcomp}
\usepackage{xcolor}
\usepackage{rotating}
\usepackage{booktabs}
\usepackage{multirow}
\usepackage{stfloats}
\usepackage[hidelinks]{hyperref}
\def\BibTeX{{\rm B\kern-.05em{\sc i\kern-.025em b}\kern-.08em
    T\kern-.1667em\lower.7ex\hbox{E}\kern-.125emX}}

\begin{document}

\title{\LARGE \bf Auto-Labelling-Based Domain Transfer for 3D Object Detection on a Bicycle-Mounted LiDAR Platform}

\author{Mario Finkbeiner$^{1}$, Max A. Buettner$^{1}$, Kanak Mazumder$^{1}$ and Fabian B. Flohr$^{1}$%
\thanks{$^{1}$The authors are with the Intelligent Vehicles Lab~(IVL), Munich University of Applied Sciences, Munich, Germany.
        {\tt\small \newline \{mario.finkbeiner, max.buettner, kanak.mazumder, fabian.flohr\}@hm.edu}}%
}

\maketitle

\begin{abstract}
Reliable 3D perception of vulnerable road users (VRUs) such as cyclists and pedestrians is essential for their safety in urban traffic and a core requirement
for autonomous driving (AD). Alongside advances in vehicle-based perception, research increasingly equips bicycles with sensors to study traffic from a perspective
native to VRUs. Such platforms still rely on LiDAR detectors originally trained on vehicle data, yet annotated 3D data from a cyclist's perspective is scarce. How well
these detectors generalise to this setting has not been evaluated. We present a 3D object detection benchmark of $1{,}027$ annotated LiDAR keyframes
(over $18{,}000$ 3D bounding boxes) from the FUSE-Bike platform in urban Munich. We evaluate four nuScenes-pre-trained detectors against $1{,}854$
human-verified ground-truth (GT) boxes both in their original form and after finetuning on training labels produced by a VRU-dedicated auto-labelling
pipeline that requires no manual annotation. The zero-shot domain gap is concentrated on the VRU classes. Finetuning recovers most of it, improving mean
average precision (mAP) by up to $23.4$ points with the largest gains on pedestrians and cyclists, and the adapted detectors even surpass the quality of the
auto-labels they were trained on. The benchmark provides a reproducible baseline for VRU-centric 3D detection and shows that auto-labels are a viable
substitute for manual annotation when adapting vehicle-trained detectors to a cyclist platform.
\end{abstract}


\section{Introduction}
Cyclists and pedestrians are among the most exposed participants in urban traffic, as they lack the passive protection of motor vehicles and depend on being
correctly perceived by the vehicles around them. Reliably detecting these vulnerable road users (VRUs) with on-board sensors is therefore a central challenge for autonomous driving (AD) in urban environments.
Alongside the continuing effort to improve this from the vehicle's perspective, bicycle-mounted research platforms have recently emerged as a tool for
observing urban traffic from within the flow of VRUs themselves, providing data and viewpoints that vehicle sensors cannot offer. Accurate 3D detection of
surrounding participants is a fundamental requirement for any such platform operating in urban traffic.

\begin{figure}[t]
    \centering
    \includegraphics[width=1.0\linewidth]{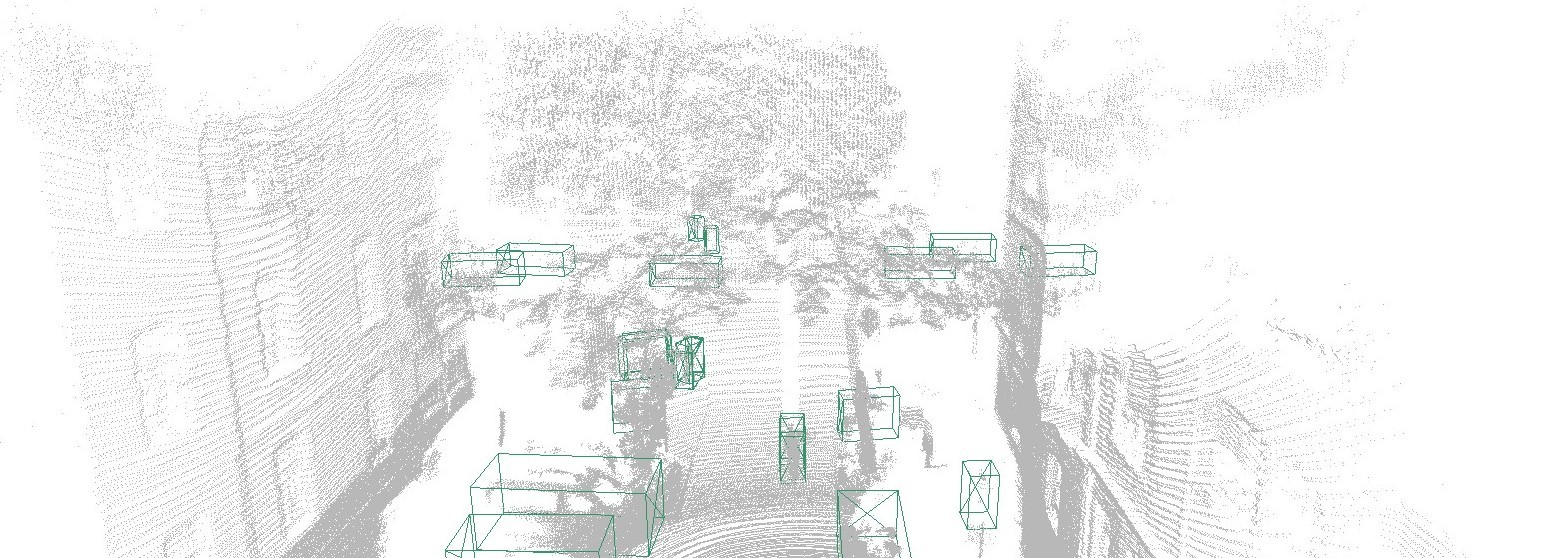}
    \vspace{0.1cm}
    \includegraphics[width=1.0\linewidth]{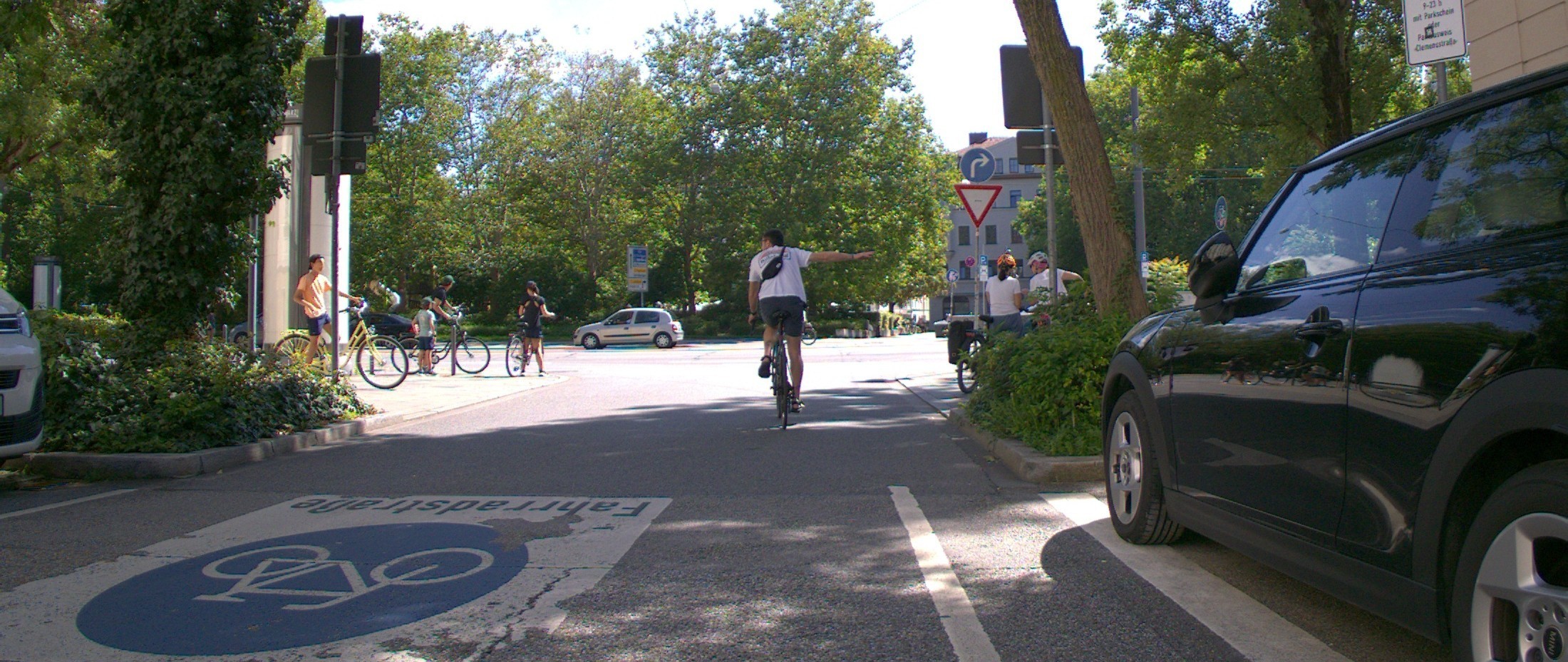}
    \caption{Example 3D auto-labels produced by the VRU-Label3D pipeline's detector ensemble on recordings from the FUSE-Bike platform in
    Munich, Germany. The vehicle-trained detectors that make up the ensemble transfer reasonably well to cars, but cyclists and pedestrians, the
    safety-critical classes on a bicycle-mounted platform, remain the main weakness.}
    \label{fig:teaser}
\end{figure}

Unlike automotive sensor suites, bicycle-mounted platforms exhibit substantial variation in sensor mounting perspective, LiDAR scan pattern,
and point density, as illustrated in Fig.~\ref{fig:teaser}. A bicycle is lower to the ground, far more agile, and surrounded by a different distribution of
road users, so detectors trained on automotive data face a significant domain shift when deployed on it. This challenge is exacerbated by the composition of
the training data itself: cyclists and pedestrians are among the rarest and most poorly annotated classes, resulting in a scarcity of training data for precisely the
classes whose safety is most at stake. Compounding this, no public benchmark currently provides multi-class VRU detection labels from a cyclist's perspective, as
manual annotation of 3D point clouds is labour-intensive and does not scale well. Without such labels, neither the performance of automotive detectors on cyclist-perspective data nor the quality of automatically generated labels as a training
signal can be rigorously assessed, even though methods that leverage large automotive datasets offer a practical path to annotation-free training data.

We tackle these problems with two main contributions:
\begin{itemize}
    \item \textbf{A cyclist-perspective VRU detection dataset} of $1{,}027$ keyframes and over $18{,}000$ 3D boxes, pairing auto-labelled training sequences from a
    VRU-dedicated pipeline with $1{,}854$ human-verified ground-truth (GT) boxes on held-out scenes, enabling both detector evaluation and auto-label quality
    assessment.
    \item \textbf{A benchmark} that quantifies the vehicle-to-cyclist domain gap across four architecturally diverse, nuScenes-pre-trained detectors and
    demonstrates that finetuning on the auto-labels alone substantially closes it, improving mAP by $13.7$ to $23.4$ points, with the
    largest gains on the safety-critical pedestrian and cyclist classes, whose per-class AP rises by up to $31.8$ points, with the finetuned
    detectors even surpassing the quality of the auto-labels they were trained on.
\end{itemize}

\section{Related Work}

\subsection{Datasets and Platforms for VRU Perception}
Robust 3D detection in AD depends heavily on annotated data, which to date has been captured almost entirely from the vehicle's perspective. Large automotive datasets
such as nuScenes~\cite{caesarNuScenesMultimodalDataset2020}, Waymo~\cite{sunScalabilityPerceptionAutonomous2020}, and Lyft~\cite{houstonOneThousandOne2021}
provide diverse urban scenes, yet observe VRUs only as they appear to a car and contain few of them, with cyclists making up less than $1\%$ of nuScenes
annotations.
Bicycle-mounted sensor platforms instead place the sensors among the VRUs themselves, offering data and viewpoints that vehicle sensors cannot. The AuRa cargo-bike~\cite{sassAuRaDatasetVision2025} and BikeScenes'
SenseBike~\cite{gorenBikeScenesOnlineLiDAR2025} target semantic segmentation, the FUSE-Bike platform underlies the BikeActions dataset~\cite{buettnerBikeActionsOpenPlatform2026} for
cyclist-perspective VRU action recognition, and SaBi3d~\cite{odenwaldSaBi3dALiDARPoint2024} addresses vehicle
detection during car-to-bicycle overtaking. None of them, however, provides multi-class 3D object detection labels for VRUs. Such labels enable a moving
platform to reason about surrounding agents in space and over time, beyond the per-point semantics of point cloud segmentation or camera-based features alone.
Therefore, they are a prerequisite for VRU-aware perception.

\subsection{3D Object Detection}
In the absence of cyclist-domain training data, perception on such platforms relies on detectors developed for automotive 3D detection, which span several
architectural paradigms. Voxel-based, anchor-based detectors such as SECOND~\cite{yanSECONDSparselyEmbedded2018} discretise the point
cloud into a regular grid processed by sparse 3D convolutions, whilst pillar- and centre-based detectors such as CenterPoint~\cite{yinCenterbased3DObject2021a}
represent objects as keypoints in a bird's-eye-view feature map. More recent designs include fully sparse, anchor-free architectures, exemplified by
VoxelNeXt~\cite{chenVoxelNeXtFullySparse2023}, and query-based transformer heads, exemplified by TransFusion~\cite{baiTransFusionRobustLiDARCamera2022}. These paradigms differ in how
they aggregate context and localise small objects, yet none has been evaluated on a cyclist platform. Because the architecture may itself shape robustness to the
viewpoint shift, characterising the domain gap calls for a representative spread of paradigms rather than a single detector.

\subsection{Domain Adaptation and Auto-Labelling}
Transferring these detectors to a cyclist platform introduces a domain shift. When no labelled target data is available, unsupervised domain adaptation (UDA)
can be used to address this issue. The resulting artefacts are typically used in one of two ways: either as finetuned models based on their own predictions
through self-training, or as pre-labels for the unlabelled domain. ST3D~\cite{yangST3DSelftrainingUnsupervised2021a} introduced self-training with random object scaling to counter
size bias between domains, CL3D~\cite{pengCL3DUnsupervisedDomain2023} addresses geometric disparities between LiDAR sensors, and
UADA3D~\cite{wozniakUADA3DUnsupervisedAdversarial2024a} aligns features adversarially. Multi-source methods such as MS3D++~\cite{tsaiMS3DEnsembleExperts2025a}
fuse predictions from an ensemble of detectors pre-trained on diverse automotive datasets, improving robustness across varying point densities, whilst
OpenBox~\cite{leeOpenBoxAnnotateAny2025} uses 2D foundation models for 3D annotation but requires multi-modal calibration that VRU platforms rarely provide.
These methods are designed as closed evaluation loops, however: their outputs do not follow a standardised dataset schema, and some even discard
under-represented classes such as cyclists, which makes them ill-suited for producing training labels that can drive established detectors and metrics. We
therefore build on VRU-Label3D~\cite{IvlVRULabel3D2026}, an extension of MS3D++ that retains the cyclist class and exports tracked auto-labels in the nuScenes
format, and use it to generate the training labels for our benchmark.

\section{The FUSE-Bike Detection Benchmark}
We construct a 3D object detection benchmark from data recorded on the FUSE-Bike platform~\cite{buettnerBikeActionsOpenPlatform2026}, a cyclist-mounted sensor suite, separating automatically labelled
training data from a held-out evaluation set with human-verified GT. Table~\ref{tab:dataset} summarises the dataset splits. As in automotive datasets,
the class distribution is dominated by vehicles, with cyclists the rarest class, reflecting natural urban traffic composition rather than a sampling choice.
The following describes the platform and recordings, the auto-labelling pipeline for training data, and the GT annotation process for the evaluation set.

\begin{table}[h]
    \centering
    \caption{FUSE-Bike detection benchmark. Training scenes use ensemble auto-labels, whereas the held-out evaluation scenes use human-verified GT.
    Training and evaluation scenes are fully disjoint, with no temporal overlap between them. Counts are reported at the 2\,Hz keyframe rate used for
    training and evaluation.}
    \label{tab:dataset}
    \begin{tabular}{lrrrrr}
        \toprule
        Split & Scenes & Frames & Vehicle & Pedestrian & Cyclist \\
        \midrule
        Train (auto-labels) & 33 & 941 & 9903 & 4311 & 2131 \\
        Test (GT) &  3 &  86 & 1093 &  396 &  365 \\
        \midrule
        Total & 36 & 1027 & 10996 & 4707 & 2496 \\
        \bottomrule
    \end{tabular}
\end{table}

\subsection{Platform and Recordings}
\label{sec:platform}
All data were recorded in urban Munich, Germany, in the vicinity of the Munich University of Applied Sciences, covering dedicated bike paths, bicycle streets,
and shared general-traffic roads, so as to capture the diversity of situations a cyclist encounters in daily commuting. The platform carries two vertically
stacked Ouster LiDARs (an OS0 and an OS2), each with 128 beams, whose point clouds are fused to combine the wide near-field coverage of the OS0 with the longer
range of the OS2. A front-facing camera, hardware-synchronised with the LiDARs, supports annotation and qualitative verification.

The continuous recordings are segmented into non-overlapping 15-second sequences, a duration chosen to balance file size against sufficient temporal context
for tracking. Ego-motion is estimated by a LiDAR-inertial odometry~\cite{malladiRobustApproachLiDARInertial2026} that fuses the LiDAR scans with inertial measurements, which handles the agile motion of a
cyclist better than a pure LiDAR-odometry baseline.

\subsection{Auto-Labelling for Training Data}

\begin{figure*}[ht]
    \centering
    \includegraphics[width=\textwidth]{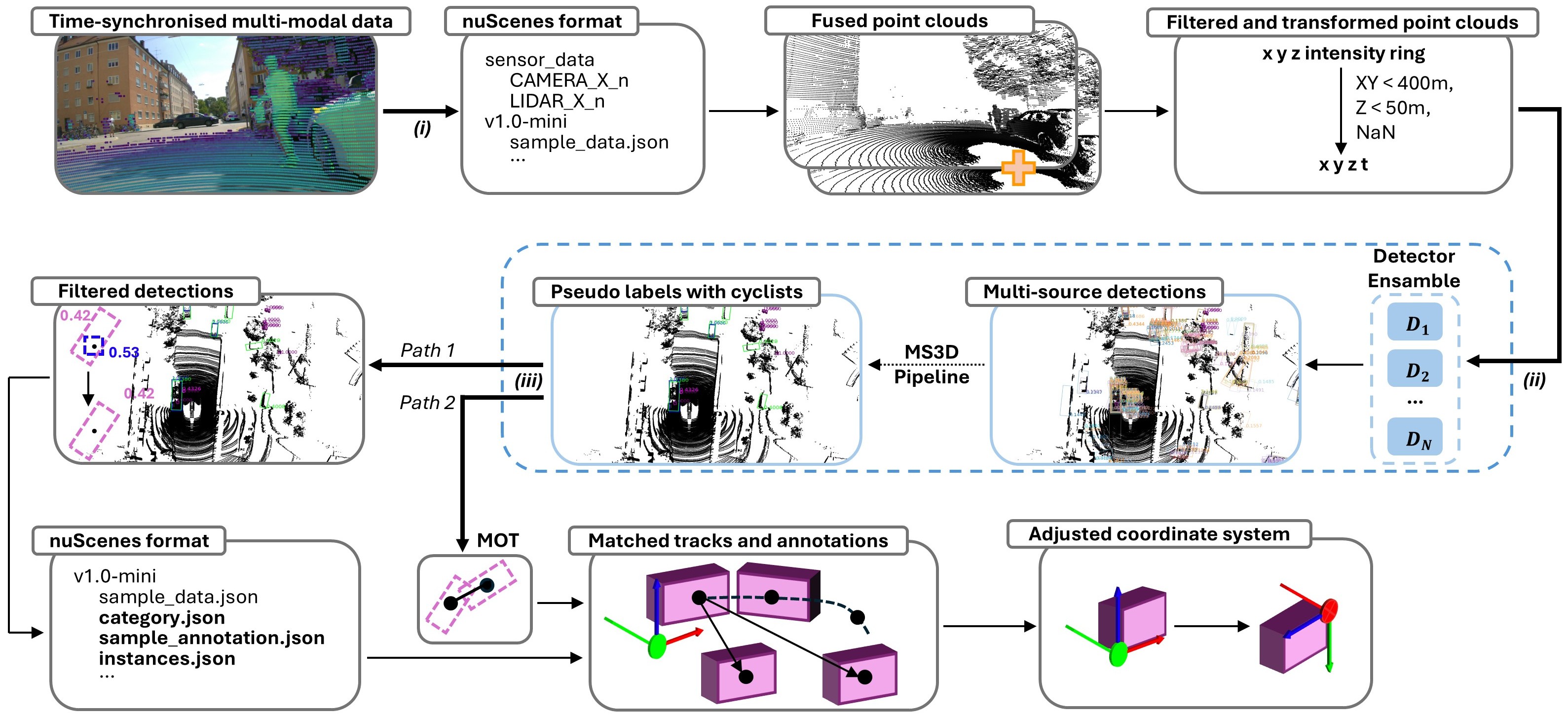}
    \caption{The VRU-Label3D auto-labelling pipeline used to generate the training labels. Time-synchronised multi-modal data (top-left) is structured into the
    nuScenes format; fused, filtered point clouds are processed by a multi-source ensemble of experts (middle); the resulting detections, including the recovered
    ``Cyclist'' class, undergo geometric filtering and multi-object tracking and are exported as standardised, tracked annotations (bottom-right).}
    \label{fig:pipeline}
\end{figure*}

\label{sec:autolabel}
Training labels are generated automatically with VRU-Label3D~\cite{IvlVRULabel3D2026}, a multi-source ensemble auto-labelling framework that extends
MS3D++ and is tailored to VRUs. As outlined in Fig.~\ref{fig:pipeline}, the pipeline ingests the synchronised LiDAR and
camera streams and runs an ensemble of detectors pre-trained on several automotive datasets (Waymo, nuScenes, Lyft). Their proposals are merged by kernel
density estimation, and the fused detections are linked across frames by a multi-object tracker to enforce temporal consistency, before being exported as
standardised nuScenes-format annotations.

Because cyclists are the scarcest and most safety-critical class, the pipeline is designed to preserve them throughout. The base MS3D++ framework discards
cyclists during its refinement stage, whereas VRU-Label3D recovers these detections and carries them through to the final labels, so that the training data
retains as many cyclists as the ensemble can recover instead of dropping them as noise. The resulting pseudo-labels are used directly as the training signal
for all benchmark detectors, without any manual correction of the training scenes.

\subsection{Ground Truth Annotation}
\label{sec:gt}
For trustworthy evaluation we manually annotate a held-out set of three sequences that are excluded from training. Annotation is performed in
SUSTechPOINTS~\cite{naurrilNaurrilSUSTechPOINTS2025}, initialised from the auto-labels so that the annotator acts as a verifier rather than a creator,
correcting box geometry and class against the synchronised camera images. The evaluation set comprises 86 labelled keyframes at 2\,Hz across the three
scenes, totalling 1854 boxes. Annotations use three classes (vehicle, pedestrian, cyclist), onto which the multi-class predictions of the pre-trained
detectors are mapped, so that all models are evaluated on the classes relevant to cyclist safety. As track identities are not required for frame-level
detection, they are not annotated in the GT.

\section{Experimental Setup}
\label{sec:setup}

\subsection{Detector Suite}
We instantiate the spread of paradigms with the four detectors listed in Table~\ref{tab:main}, all taken from OpenPCDet~\cite{OpenmmlabOpenPCDet2026a} with their
publicly released nuScenes-pre-trained checkpoints. Holding the source domain fixed in this way ensures that differences in transfer reflect the architecture
rather than the training data. Concretely, we use CenterPoint with a dynamic-pillar encoder, SECOND with an anchor-based multi-head detector (SECOND-MH), the
LiDAR-only TransFusion-L, and the fully sparse VoxelNeXt. Evaluating this range lets us test whether the domain gap and the benefit of finetuning are consistent
across detector families or specific to a single design.

\begin{table*}[ht]
    \centering
    \caption{Benchmark results on the held-out GT test set: nuScenes-style mAP and per-class AP (\%), zero-shot vs.\ finetuned on the auto-labels.
    Per-class AP is the mean over the $\{0.5,1,2,4\}$\,m centre-distance thresholds. The gain column is the relative mAP improvement from finetuning.
    Best per column in \textbf{bold}.}
    \label{tab:main}
    \begin{tabular}{lccccccccc}
        \toprule
        & \multicolumn{4}{c}{Zero-shot (nuScenes pre-trained)} & \multicolumn{4}{c}{Finetuned (FUSE-Bike auto-labels)} & \multirow{2}{*}{mAP gain (\%)} \\
        \cmidrule(lr){2-5} \cmidrule(lr){6-9}
        Detector & mAP & Vehicle & Pedestrian & Cyclist & mAP & Vehicle & Pedestrian & Cyclist & \\
        \midrule
        CenterPoint~\cite{yinCenterbased3DObject2021a}   & 48.4 & 68.5 & 27.9 & 48.9 & 71.8 & 81.3 & 59.7 & 74.4 & \textbf{+48.3} \\
        SECOND-MH~\cite{yanSECONDSparselyEmbedded2018}     & 56.6 & 78.6 & 44.1 & 47.0 & 74.2 & 80.7 & 68.3 & 73.5 & +31.1 \\
        TransFusion-L~\cite{baiTransFusionRobustLiDARCamera2022} & 62.3 & 78.8 & 49.7 & 58.4 & 77.1 & \textbf{83.8} & 69.8 & 77.7 & +23.8 \\
        VoxelNeXt~\cite{chenVoxelNeXtFullySparse2023}     & \textbf{63.5} & \textbf{79.3} & \textbf{51.8} & \textbf{59.5} & \textbf{77.2} & 83.5 & \textbf{70.5} & \textbf{77.7} & +21.6 \\
        \bottomrule
    \end{tabular}
\end{table*}

\subsection{Finetuning Protocol}
Each detector is evaluated in two settings. In the \textit{zero-shot} setting we run the publicly released nuScenes-pre-trained checkpoint without any
adaptation, which exposes the domain gap. In the \textit{finetuned} setting we adapt the same checkpoint on the FUSE-Bike training auto-labels for 20 epochs.
The detection head is re-grouped from the ten nuScenes classes to the three target classes (car, pedestrian, bicycle), so that the backbone and shared weights
transfer from the pre-trained model whilst the re-grouped head layers are reinitialised. The training data, schedule, and detection range are kept identical
across detectors to isolate the effect of architecture. The detection and evaluation range is clamped to $\pm 40$\,m, beyond which the fused OS0/OS2 returns
are too sparse for reliable boxes.

\subsection{Evaluation Metrics}
We follow the nuScenes detection protocol, matching predictions to GT by 2D centre distance rather than by IoU, the latter being unstable for the
sparse, partially observed returns of a low-mounted sensor. For each class $c$ and centre-distance threshold $d\in\mathcal{D}=\{0.5, 1, 2, 4\}$\,m, predictions
are matched to GT within $d$. Our primary metric is the mean over the thresholds and the three classes $\mathcal{C}$,
\begin{equation}
    \mathrm{mAP} = \frac{1}{|\mathcal{C}|\,|\mathcal{D}|} \sum_{c\in\mathcal{C}} \sum_{d\in\mathcal{D}} \mathrm{AP}_{c,d}.
    \label{eq:map}
\end{equation}
To characterise behaviour at a concrete operating point relevant to deployment, we additionally report the true positive (TP), false positive (FP) and false negative
(FN) counts at a fixed threshold (score $\geq 0.3$, match $\leq 2$\,m), and the precision and recall they induce,
\begin{equation}
    \mathrm{Precision} = \frac{\mathrm{TP}}{\mathrm{TP}+\mathrm{FP}}, \qquad
    \mathrm{Recall} = \frac{\mathrm{TP}}{\mathrm{TP}+\mathrm{FN}},
    \label{eq:pr}
\end{equation}
the same quantities whose threshold-swept curve defines $\mathrm{AP}_{c,d}$ above. For VRUs a missed detection carries a higher safety cost than a spurious one,
making recall the more critical of the two. The pre-trained detectors predict the full nuScenes label set, which is mapped to the three target
classes so that the zero-shot and finetuned settings are directly comparable.

\section{Results}

\subsection{Quantifying the Domain Gap}
The left half of Table~\ref{tab:main} reports zero-shot performance, and the pattern is consistent across all four detectors. Vehicles transfer reasonably
well, whereas the two VRU classes are markedly weaker. Within every detector the vehicle class outperforms the worst VRU class by a wide margin, from
$27.5$ points for the strongest baseline to $40.6$ points for the weakest. The domain gap is therefore not uniform but concentrated on the safety-critical
classes, exactly those under-represented in the automotive pre-training data.

Overall mAP also varies widely, from $48.4\%$ for CenterPoint to $63.5\%$ for VoxelNeXt. The more recent sparse-voxel and transformer designs transfer better
than the pillar-based CenterPoint and the older SECOND, showing that the architecture itself affects robustness to the viewpoint shift. The low sensor mounting
exacerbates this issue, as the recall-oriented detectors identify many false positives in the form of pedestrians from clusters of points on poles, vegetation
and street infrastructure, which resemble standing people far more than they do in elevated automotive scans.

\begin{figure}[ht]
    \centering
    \includegraphics[width=\columnwidth]{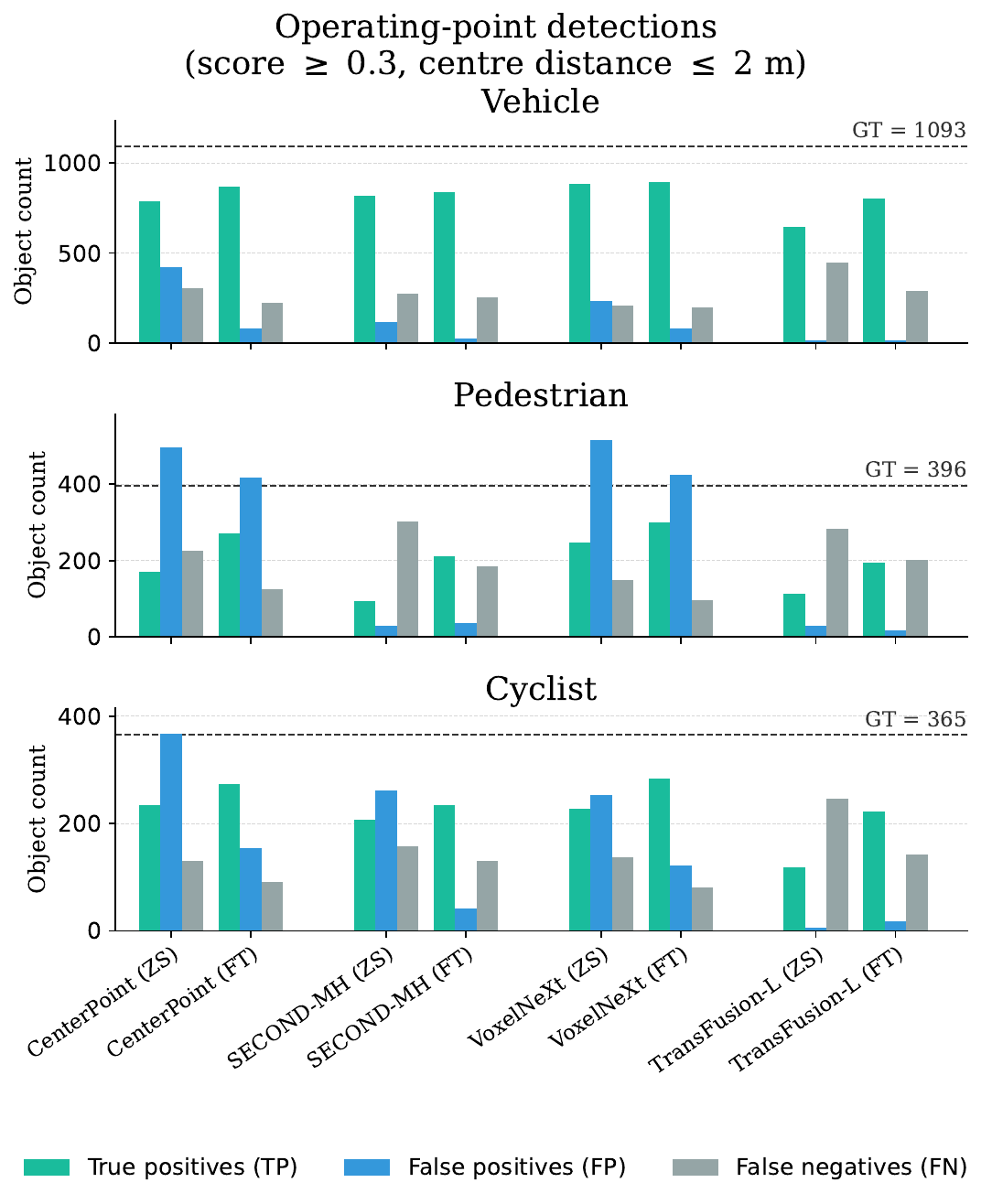}
    \caption{Operating-point detection counts (score $\geq 0.3$, centre distance $\leq 2$\,m) per class. ZS\,=\,zero-shot, FT\,=\,finetuned. For each detector
    the two settings are placed side by side, with three bars each: true positives (TP), false positives (FP) and false negatives (FN). The dashed line marks the
    GT total per class (TP\,$+$\,FN).}
    \label{fig:perclass}
\end{figure}

\subsection{Adapting Detectors via Auto-Labels}
The right half of Table~\ref{tab:main} shows that finetuning on the auto-labels improves every detector, despite the complete absence of manually labelled
training data. The gains in mAP reach up to $23.4$ points, and just as importantly the spread between detectors shrinks from $15.1$ points in the zero-shot
setting to $5.4$ points afterwards. Adaptation thus largely removes the architecture-dependent part of the gap and brings the four detectors to a comparable level.

The improvement is driven by the VRU classes, whose AP rises by $18.2$ to $31.8$ points, whereas the already-strong vehicle class gains only modestly. In
relative terms the gains land exactly where a cyclist platform needs them: pedestrian AP more than doubles for CenterPoint ($+114\%$) and grows by a third
even for the strongest baseline, and cyclist AP rises by up to $56\%$. These are precisely the classes that automotive datasets under-represent, so the
benefit of adaptation is largest on the participants most central to cyclist safety.
Fig.~\ref{fig:perclass} makes the underlying mechanism explicit at the operating point. For every detector finetuning turns missed objects into true positives and
raises recall on all three classes, and where the zero-shot detectors over-predicted VRUs it also removes most of those false positives. A class-dependent trade-off
remains, however. The anchor- and query-based detectors
favour precision on pedestrians, producing few false positives but lower recall, whereas the centre-based and sparse detectors favour recall at the cost of more
false positives. Pedestrian stays the hardest class throughout and never reaches the level of the vehicle class, which we again attribute to the low sensor mounting.

\subsection{Qualitative Results}
\begin{figure}[t]
    \centering
    \includegraphics[width=\columnwidth]{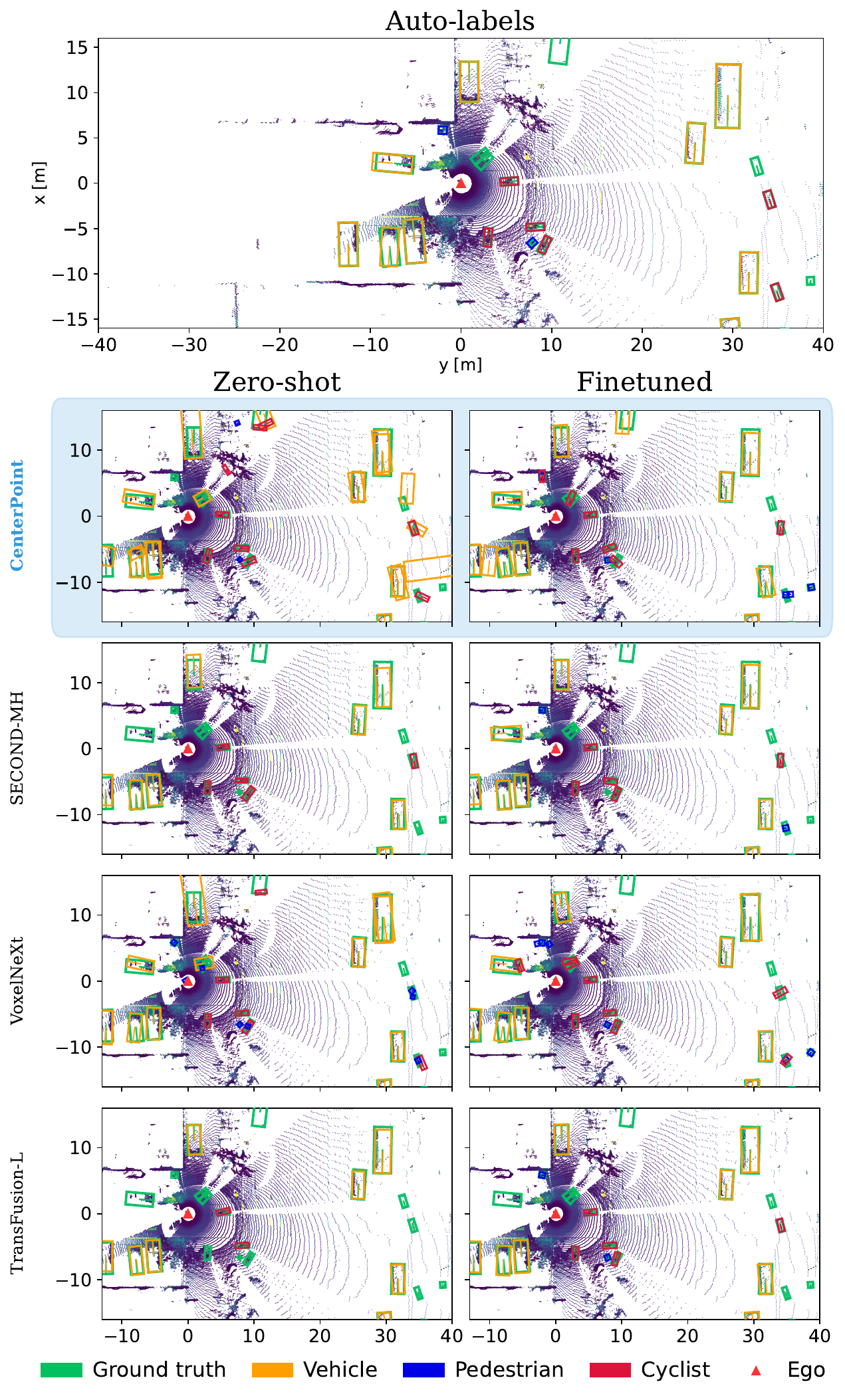}
    \caption{BEV detections for one held-out frame (10 vehicles, 9 cyclists, 3 pedestrians). Top: the ensemble auto-labels; below, each row is one detector,
    zero-shot (left) vs.\ finetuned (right), as in Table~\ref{tab:main}. The highlighted CenterPoint row gains the most from finetuning. Boxes follow the nuScenes
    colours (yellow = vehicle, blue = pedestrian, crimson = cyclist) over the human GT (green); points are coloured by intensity.}
    \label{fig:qualitative}
\end{figure}

Fig.~\ref{fig:qualitative} contrasts zero-shot and finetuned detections on a representative held-out frame.
Before adaptation, the detectors frequently miss cyclists or split a single VRU into spurious detections of several classes. After finetuning these cases are
resolved, with cyclists and pedestrians detected as single, correctly classified objects.

\subsection{Auto-Label Quality}
Since the auto-labels are the only training signal, we assess their quality directly by evaluating them against the human GT, treating the pipeline
as a detector. As Table~\ref{tab:prelabel} shows, the auto-labels reach $63.9\%$ mAP, on par with the best zero-shot detector. Their main limitation is recall:
the conservative ensemble misses objects, cyclists most of all.

The notable result is that all four finetuned detectors exceed the quality of the auto-labels they were trained on, in mAP and on each VRU class individually.
The detectors therefore do not merely reproduce their noisy training labels but generalise beyond them, recovering objects the labels missed. This shows that
imperfect auto-labels are a sufficient training signal for adapting vehicle-trained detectors to the cyclist domain.

\begin{table}[ht]
    \centering
    \caption{Quality of the ensemble auto-labels measured directly against the human GT on the held-out scenes (nuScenes-style AP, \%).
    All four finetuned detectors in Table~\ref{tab:main} exceed these values.}
    \label{tab:prelabel}
    \begin{tabular}{lcccc}
        \toprule
        & mAP & Vehicle & Pedestrian & Cyclist \\
        \midrule
        Auto-labels vs.\ GT & 63.9 & 79.6 & 51.7 & 60.2 \\
        \bottomrule
    \end{tabular}
\end{table}

\subsection{Labelling Effort}
The practical benefit of the approach is that the entire $941$-frame training set was labelled automatically, with manual annotation required only for the
$86$-frame evaluation set. Adapting a vehicle-trained detector to the cyclist platform therefore incurs essentially no annotation cost, and manual effort is
reserved for the small, high-quality test set needed for trustworthy evaluation. This makes the approach practical for scaling to new recordings and
platforms, where collecting data is cheap but labelling it is not.

\section{Conclusion}
We introduced a cyclist-perspective 3D object detection benchmark, comprising both a multi-class VRU dataset and a suite of four adapted detectors. The dataset pairs
$941$ auto-labelled training frames with a compact, hand-labelled test set of $86$ keyframes, together over $18{,}000$ 3D annotations from a viewpoint no
existing benchmark covers, of which $1{,}854$ are human-verified. Using it, we quantified the vehicle-to-cyclist domain gap across four architecturally diverse,
nuScenes-pre-trained detectors and found it consistent across architectures and concentrated on the safety-critical VRU classes: vehicles transfer reasonably
well, whereas pedestrians and cyclists degrade sharply in the zero-shot setting. Finetuning each detector on auto-labels generated without any manual annotation
improves mAP by $13.7$ to $23.4$ points ($21.6$ to $48.3\%$ relative) and largely removes the architecture-dependent component of the gap. Crucially, the largest gains
fall on the pedestrian and cyclist classes that automotive datasets under-represent but a cyclist platform must perceive, with pedestrian AP more than doubling
for the weakest baseline. The finetuned detectors even surpass the quality of the auto-labels they were trained on, confirming that imperfect auto-labels are a
sufficient signal for closing the vehicle-to-cyclist domain gap. Together these results provide a reproducible baseline for VRU-centric 3D detection
and establish auto-labelling-based domain transfer as a practical, annotation-free way to bring vehicle-trained detectors to new platforms.\\

\bibliographystyle{IEEEtran}
\bibliography{references}

\end{document}